%% file: lamkin.tex
\documentclass{article}

\usepackage{scml2025}

\usepackage{james}

\usepackage{multicol}
\usepackage[utf8]{inputenc} 
\usepackage[T1]{fontenc}    
\usepackage{hyperref}       
\usepackage{url}            
\usepackage{booktabs}       
\usepackage{amsfonts}       
\usepackage{nicefrac}       
\usepackage[stretch=35,shrink=35]{microtype}      
\usepackage{cleveref}       
\usepackage{lipsum}         
\usepackage{graphicx}
\usepackage{natbib}
\usepackage{doi}
\usepackage{wrapfig}
\usepackage{float}
\usepackage{lipsum}
\usepackage{alltt}          
\AtBeginDocument{\lefthyphenmin=3 \righthyphenmin=3 \hyphenpenalty=1000 }
\renewcommand{\pos}{\ensuremath{\xi}} 

\title{Improving regional weather forecasts with neural interpolation}
\renewcommand{\shorttitle}{Improving regional weather forecasts}

\date{}

\newif\ifuniqueAffiliation
\uniqueAffiliationtrue

\ifuniqueAffiliation 
    \author{%
      James Jackaman \\
      Norwegian University of Science and Technology \\
      \texttt{james.jackaman@ntnu.no} \\
      \And
      Oliver J. Sutton \\
      King’s College London and Synoptix Ltd \\
      \texttt{oliver.sutton@kcl.ac.uk} \\
    }
  \else
    \usepackage{authblk}
    
    \author[1]{%
      James Jackaman\thanks{\texttt{james.jackaman@ntnu.no}}}%
    \affil[1]{Department of Mathematics and Statistics,
      NTNU, 7491 Trondheim, Norway}
  \fi

\begin{document}
  \maketitle
  \vspace{-1.1cm}
  \begin{abstract}
    In this paper we design a neural interpolation operator to
    improve the boundary data for regional weather models, which is a
    challenging problem as we are required to map multi-scale dynamics
    between grid resolutions. In particular, we expose a methodology
    for approaching the problem through the study of a simplified
    model, with a view to generalise the results in this work to the
    dynamical core of regional weather models. Our approach will
    exploit a combination of techniques from image super-resolution
    with convolutional neural networks \emph{(CNNs)} and residual
    networks, in addition to building the flow of atmospheric dynamics
    into the neural network. We find that building in the flow
    more than halves the squared error of small networks.
  \end{abstract}


  \begin{multicols}{2}

    \section{Motivation}

    In recent years, machine learning has revolutionised many areas of
    scientific computing. One area of recent significant advancement
    is weather prediction. Due to its multi-scale nature and the
    chaotic nature of the underlying dynamics, weather forecasting has
    long employed statistical tools. Indeed, to improve the accuracy
    of forecasts, models are enhanced through the incorporation of
    observations (known as data assimilation) \cite{Navon, Rabier2005,
      Fisher2008}. Additionally, unresolvable phenomena (such as cloud
    coverage and precipitation) are often approximated by statistical
    models \cite{Berliner2003, Straka2009}.

    Machine learning weather forecasting models have recently been
    developed through graph neural networks by Google DeepMind,
    and have matured to the extent that they can
    compete with operational forecasts over 10 days
    \cite{Lam2023}. One limitation of such models is that they do
    not take careful consideration of physical laws, although
    significant progress is being made in this direction,
    \cite[e.g.]{CardosoBihlo2025, David2023, Jagtap2020,
      Celledoni2023}, and the field is rapidly developing. Our
    philosophy behind this work is to marry the recent progress in
    machine learning with traditional approaches taken by
    meteorological centres.
    
    Due to the chaotic and multi-scale nature of atmospheric dynamics,
    to accurately simulate weather locally one must resolve the problem
    to a high resolution. While the method of simulation has
    changed over time and varies between meteorological institutes
    \cite{Ullrich2017, Melvin2024}, existing models have
    invariably been derived from numerical discretisations of PDEs. In
    practice, these models reduce to solving a (non)linear system of
    equations, which for a high resolution model becomes prohibitively
    (computationally) expensive. A key tool in improving this
    is
    parallelisable iterative linear solvers \cite{Maynard2020, Muller2014, Wesseling2001}, which
    have recently been implemented for Navier--Stokes
    \cite{Danieli2022, self:rewaves, Lohmann2023}. To avoid prohibitive computational costs, one
    approach meteorologists take is to restrict their model to a region
    of high interest and run this much smaller model at a high
    resolution \cite{Bush2020, Bush2023}. Unfortunately, this approach
    has a major limitation. Information must both enter and leave the
    region in a physically meaningful (and qualitatively accurate)
    manner. This has long been the bane of regional weather models, and
    is a prime area for enhancement by neural networks. Here,
    we shall refer to regional weather models as limited area models
    \emph{(LAMs)}.
    
    Typically, fine scale regional weather models are driven by a
    coarse global model which informs the boundary data. A large
    amount of nonphysical behaviour appears near the boundaries of the
    LAM due to differences in physics at different scales. While not
    typical in meteorological models, it is possible to decrease this
    effect by grading the mesh to slowly increase resolution on the
    boundary. Unfortunately, this approach remains unable to resolve
    some fine-grain phenomena (such as eddies and vortices) moving
    onto the LAM, as they do not appear on the coarse global
    model. One approach often taken at meteorological centres can be
    described as follows: Around the area of interest, one may
    introduce an interface (or blending) region, as shown in Figure
    \ref{fig:mesh}.  This region varies in size, but typically
    requires an area similar to that of the LAM. Initially, the
    solution in the interface region and LAM is interpolated directly
    from the coarse global model. The model is then run over time, and
    at every time step, the solutions on the global model and LAM are
    averaged in the blending region. After a certain amount of time,
    the dynamics in the LAM stabilise and appear physically
    meaningful. This is known as spinning up the model.

    \begin{figure}[H]
      \caption{A visualisation of the coarse global model, LAM (regional
        model) and interface region.}
    \begin{center}
      \input{figures/mesh.tex}
    \end{center}
    \label{fig:mesh}
  \end{figure}
    
  In this work, \emph{we aim to improve the spinning up of the model
    by replacing the averaging between the global model and LAM with a
    neural interpolation operator}. This interpolation aims not only
  to accurately carry information from the coarse grid to the fine
  grid, but also to infer unresolvable features of the fine grid from
  high resolution training data. Our approach will utilise techniques from
  image super-resolution \cite{Dong2016, Wang2021} and residual neural
  networks \cite{He2016, Celledoni2021}, and will also
  incorporate the flow dynamics over time. To
  clarify the exposition (as well as highlighting potential pitfalls) we
  consider a significantly simplified model with the aim of generalising
  what we learn in our simplified setting to more complex fluid
  dynamics models. In fact, for the approach we outline, the main
  difficulty in generalising lies in data representation and not in
  simulating the appropriate physics. This will be the subject of future work.

  \section{Methodology} \label{sec:methodology}

  Atmospheric dynamical cores are typically driven by the Euler
  equations in an ideal gas. Here, we consider the simplified model of
  the shallow water equation \emph{(SWE)}, as this is often the first
  case study considered when developing new dynamical cores,
  \cite[e.g.]{Staniforth2013, Kent2023}. The SWE describes a shallow
  layer of fluid in hydrostatic balance bounded from below by bottom
  topography and above by a free surface. In two dimensional space,
  the equation is described by two variables, namely velocity in the
  $(x,y)$-direction and pressure. Similarly to \cite{Staniforth2013},
  to further simplify we make the assumption that our solution is
  constant in $y$, reducing the problem as follows: Let $x \in S^1$
  be periodic and $t\in[0,T]$ for some fixed end time $T$, then the
  SWE is
  \begin{equation} \label{eqn:swe}
    \begin{split}
      u_t - f v + p_x = 0,
      & \quad
        v_t + f u = 0, \\
      p_t + g u_x & = 0,
    \end{split}
  \end{equation}
  where $u=u(t,x),v=v(t,x)$ are velocities in the $x$- and $y$-directions,
  respectively, and $p=p(t,x)$ is the pressure. Further, $f$
  and $g$ are positive constants where $f$ represents the Coriolis
  parameter and $g$ an initial reference pressure value. We note that,
  even though our model assumes functions are constant in $y$, the
  velocity in the $y$-direction is not constant due to the Coriolis
  parameter. While vortices and eddies are not captured by this simplified model, mapping
  solutions between scales still represents a significant
  challenge. The SWE supports three key types of waves: gravity waves,
  inertial waves and travelling waves. High frequency travelling waves
  are particularly difficult to model in the discrete setting. Beyond
  a certain frequency these are known to behave nonphysically and (on
  sufficiently coarse meshes) alias to lower frequency waves, see
  \cite[e.g.]{Melvin2018}.  In \S\ref{sec:implementation} we shall
  consider an amalgamation of these types of waves to maximise the
  expressivity of our neural interpolant.

  To maximise compatibility between models, we utilise the same
  discretisations for the global model and LAM. In particular,
  we use a compatible finite element discretisation \cite{Cotter2023,
    Cotter2014} reduced to $1$D space. We outline this discretisation
  in more detail in Appendix \ref{a:fe}, including the temporal
  discretisation. After partitioning our periodic spatial domain, our
  discretisation is given by \emph{piecewise polynomials} within each
  element of the domain. In particular, each component of our velocity
  space is described by a \emph{continuous piecewise linear
    function}, and the pressure by a \emph{discontinuous piecewise
    constant function}. Our velocity space can be uniquely
  characterised by the values of the solution at nodal points, and
  conversely the pressure by the value of the solution at the midpoint
  of each element, as visualised in Figure \ref{fig:fes}. The points
  at which we evaluate the solution are known as degrees of freedom
  \emph{(DOF)}. Our velocity and pressure may be represented as
  vectors comprised of the DOF in our approximation, and these vectors
  form the
  data structures we need to process with our neural network. Before
  proceeding, we make two important remarks: Firstly, that the
  velocity space and pressure space are fundamentally different and should
  be treated as such. Secondly, that the DOF are equidistributed,
  which allows us to act directly on the DOF with a CNN.

  \vspace{-0.3cm}
  \begin{figure}[H]
    \centering
    \caption{A coarse visualisation of the 1D mesh with three nodes
      and two elements. We represent a function in the
      velocity space in blue, marking the DOF with crosses, and a
      function in the pressure space in red, again marking the DOF
      with crosses. Note that, as the domain is periodic,
      $u(0)=u(3)$. \label{fig:fes}}
    \input{figures/fes.tex}
  \end{figure}
  \vspace{-0.3cm}

  For clarity of exposition, here we envision our entire domain as the
  blending region of the LAM. We consider both a coarse
  ``global'' model over the periodic interval $[0,3)$ and the
  ``regional'' model over the same interval. In this case study, our
  models will simply differ in resolution, with the LAM having
  \emph{four times} the resolution of the coarse global
  model. We fix the temporal resolution of both models to be the same,
  although we note that this is not fundamental to our methodology.

  Our goal is to take the DOF of a solution simulated on the coarse
  scale and interpolate to the DOF of a solution on the fine
  scale. Our approach is to initialise both the coarse and fine models
  with highly oscillatory data, and as such, the dynamics will evolve
  at different speeds on different meshes. We aim to learn not only
  how to interpolate the initial conditions onto a finer mesh, but
  also the difference in how the dynamics evolve on the different
  meshes with a neural interpolator. Here, our methodology takes
  inspiration from image super-resolution \cite{Dong2016}: We begin by
  interpolating our coarse data onto the fine mesh data structure,
  which is an exact operation if our spatial meshes are nested. After
  interpolating onto the fine mesh, we label this data as $\vec{x}_c$,
  which will be the input data for our neural network. Here
  $\vec{x}_c \in \mathbb{R}^{N \times D}$ where $N$ is our number of
  time levels and $D$ the total dimension of the spatial problem on the
  fine mesh. This neural network will be trained against the same
  initial data which has evolved natively on the fine mesh, denoted
  $\vec{x}_f \in \mathbb{R}^{N \times D}$. The core of our model is a
  CNN, chosen largely for its relatively small kernel size and
  ability to capture and incorporate derivative information \cite{Pratt2006,
    Dong2017, self:pixels}, and is structured similarly to a U-Net
  \cite{Long2015, Siddique2021}. In practice, here we will stack time
  into the channel dimension and construct U-Net-like CNNs for each
  component in the velocity and pressure independently. This choice is
  made due to the differences in data structure between velocity and
  pressure. Following PyTorch syntax, this leads to a neural network
  of the type
\begin{verbatim}
def UNet(x):
  skip = x
  x = ReLU(Conv1d(N, s1, kernel_size=3)(x))
  x = ReLU(Conv1d(s1, s2, kernel_size=5)(x)) 
  x = ReLU(Conv1d(s2, s2, kernel_size=7)(x))
  x = ReLU(Conv1d(s2, s1, kernel_size=5)(x))
  x = Conv1d(s1, N, kernel_size=3)(x) + skip
  return x
\end{verbatim}
  where \verb|x| may be $\vec{u}_c$, $\vec{v}_c$ or $\vec{p}_c$ with
  $\vec{x}_c =: \bs{\vec{u}_c,\vec{v}_c,\vec{p}_c}^T$. Here, we use three such networks, which do
    not share weights, and define $\operatorname{UNet}(\vec{x}_c) :=
    [\operatorname{UNet}_u(\vec{u}_c),\allowbreak \operatorname{UNet}_v(\vec{v}_c),\allowbreak
      \operatorname{UNet}_p(\vec{p}_c)]^T$. We note that the skip connection here is crucial. For
  example, if we aim to learn the identity operator, this is a
  surprisingly difficult task. However, through the inclusion of a
  skip connection we may capture the identity mapping through choosing
  all parameters to be zero. For our problem, the inclusion of a skip
  connection means the neural network only needs to learn the error
  between the desired output and the identity operator.

  As we understand how the dynamics evolve on the fine scale over
  time, it may be possible to improve upon our network. In particular, we
  aim to give our network the ability to learn the
  difference between dynamics at different scales. Fortunately, our
  problem is linear, so it is possible to find the flow map of the
  fine dynamics over time, which is a matrix operator
  $A\in\mathbb{R}^{D\times D}$ mapping the solution between time
  levels following the fine dynamics. With this operator in mind, we
  modify our network by
\begin{alltt}
def LearnFlow(x):
  for i in range(1, N):
    x[i,:] = A^i x[0,:]
  x = UNet2(x)
  return x
\end{alltt}
  where $\texttt{x}=\operatorname{UNet}(\vec{x}_c)$ and \verb|UNet2| does not share weights
  with the previously introduced \verb|UNet| but for simplicity has
  the same structure, excluding the skip connection. Astute readers may question the validity of
  utilising the operator $A$. We believe this is a reasonable object
  to include in our neural network, since in the full problem we will only
  utilise it in the blending region, and it will have a complexity on
  the order of magnitude of the LAM. For more complex dynamics we may
  employ cheaper approximations inspired by iterative (non)linear
  solvers. With these functions in mind, we define our neural
  interpolant as
  \begin{equation} \label{eqn:nn}
    \NN(\vec{x}_c) :=
    \operatorname{UNet}(\vec{x}_c)
    + \operatorname{LearnFlow}(\operatorname{UNet}(\vec{x}_c))
    .
  \end{equation}
  That is to say, we evolve the first time level of
    $\operatorname{UNet}(\vec{x}_c)$ by the fine dynamics, from which
    \texttt{UNet2} learns a correction to each component (see Figure \ref{fig:network} in Appendix \ref{a:training}).
    As \texttt{UNet2} has no skip connection, \eqref{eqn:nn} is the
    identity when all parameters are zero.

  Before discussing the implementation of \eqref{eqn:nn}, we briefly
  remark upon some useful properties of the SWE. The SWE
  \eqref{eqn:swe} is a \emph{conservative} PDE. A prime example of
  this is the conservation of energy over time. In the continuous
  setting, this conservation can be expressed as
  \begin{equation} \label{eqn:energy}
    \ddt \int_{S^1} \bc{g \bs{
        u^2 + v^2}
      + p^2} \di{x}
    =
    0
    .
  \end{equation}
  This is a powerful result, as \eqref{eqn:energy} immediately
  provides stability of the solution over time in the spatial $L_2$
  norm. In finite element analysis, when mapping solutions between
  mesh resolutions, it is typical to utilise a projection operator
  which is stable in an appropriate norm. To be stable in terms of
  \eqref{eqn:energy} one may choose the $L_2$ norm. Ideally, this
  property is something we would like to mimic in our neural
  interpolant \eqref{eqn:nn}. Unfortunately, the value of the discrete
  energy is dependent on the mesh resolution. The
  discrete energies on the coarse and fine grids will not be the same,
  and we cannot know the correct energy on the fine grid without
  knowing the fine solution. This prohibits us from being able to
  exactly enforce conservation of energy in our neural interpolator,
  which in turn would guarantee stability of the operator. We can,
  however, improve stability through weakly enforcing that the
  \emph{coarse} energy is preserved. That is to say, we can compute
  the initial energy of the input $\vec{x}_c$ and penalise the energy
  of the output to be approximately the same. This is physically
  meaningful (assuming the energies do not vary significantly between
  grids) and adds a form of $L_2$ regularisation to the optimisation
  problem.

  \section{Implementation} \label{sec:implementation}

  We will now discuss the implementation of our methodology. Our
  implementation can be found at \cite{self:code}, and depends on
  PyTorch 2.1.2 \cite{PyTorch}. Experiments are run on an Apple M2 Pro
  with $16$\,GB of RAM. The training data was generated with Firedrake
  \cite{Firedrake}. All experiments in this section are driven by the
  finite element method described in Appendix \ref{a:fe}, where we fix
  our coarse mesh to have an element size of $0.04$ compared to a fine
  mesh element size of $0.01$. Throughout, we fix our time step to be
  $0.01$. Further, we choose the parameters of \eqref{eqn:swe} to be
  $f=0.1$ and $g=1$. Our training data is generated by the initial conditions
  \begin{equation} \label{eqn:data}
    \begin{split}
      & \qquad  u(0,x)  = 0, \quad  v(0,x) = 0, \\
      p(0,x) & = \alpha \bc{
               e^{-\beta (x - \pos)^2}
               + \frac{1}{10} \sin{2\pi k (x - \pos)}}
      ,
    \end{split}
  \end{equation}
  where $\alpha \sim \mathcal{U}(\frac12,2)$,
  $\beta \sim \mathcal{N}(100,6^{2})$,
  $\pos \sim \mathcal{U}(1,2)$
  and $k \sim \mathcal{U}^{\operatorname{int}}(4,10)$, with
  $\mathcal{U}$, $\mathcal{U}^{\operatorname{int}}$ and $\mathcal{N}$ denoting the
  uniform, uniform integer and normal distributions, respectively. Such
  waves are an amalgamation of inertia-gravity and travelling
  waves. We generate $1000$ runs and utilise $N=10$ time levels for
  training. We use $70\%$ of our data for training and the remainder
  for validation.

  We build the neural interpolator \eqref{eqn:nn} by fixing the \texttt{UNet}
  sizes to be $s_1 = 2^3 \cdot N$ and $s_2 = 2^6 \cdot N$. To train our network, we use a batch size of $16$ and
  optimise against the square of the $L_2$ error in the finite element
  space. We use the Adam optimiser and train over $300$ epochs, as described in Appendix \ref{a:training}.

  As discussed in \S\ref{sec:methodology}, we shall also study the
  weak enforcement of energy conservation, as this gives a weak notion of stability
  to our network in addition to regularising the network,
    where we weight the energy penalty in the loss function by a fixed
    penalty parameter $\sigma$ (see Appendix \ref{a:training}).

  We run simulations with no regularisation ($\sigma=0$), and varying
  levels of regularisation, and in Table \ref{tab:loss} report the
  square of the $L_2$ error
    on the validation data.
  \vspace{-0.2cm}
  \begin{table}[H]
    \caption{The square of the $L_2$
        error on the validation data for various values of
      regularisation, as the mean $\pm$ standard deviation
        over three random seeds. \label{tab:loss}}
    \centering
    \renewcommand{\arraystretch}{1.2}
    \begin{tabular}{|l|l|}
      \hhline{|-|-|}
      $\sigma$ & Error \\
      \hhline{|-|-|}
      $0$ & $(3.94 \pm 0.11) \cdot 10^{-8}$ \\
      $0.1$ & $(1.98 \pm 0.03) \cdot 10^{-7}$ \\
      $1$ & $(1.92 \pm 0.01) \cdot 10^{-6}$ \\
      $10$ & $(6.33 \pm 0.66) \cdot 10^{-5}$ \\
      \hhline{|-|-|}
    \end{tabular}
  \end{table}
  \vspace{-0.2cm}
  We observe that, with no regularisation, the
    error is of the order of $10^{-8}$.
  We notice that, as we
  increase the regularisation, the error increases. In fact, once $\sigma=10$ our model becomes
  significantly less accurate. For small values of $\sigma$, the error
  remains competitive.
  We note that the penalty
    slightly improves the conservation of energy (see Appendix
    \ref{a:training}). We further note that
  the $L_2$ errors are
  small for $\sigma \leq 1$.

\begin{figure}[H]
    \caption{The square of the $L_2$ error on the validation
        dataset with $\sigma = 0$ against the number of convolutions
        $L$ in each \texttt{UNet}, for $s_1 = 2^3 \cdot N$, where we
        shade the gain from the flow map. On the right, ``full''
        uses the \texttt{UNet} of \S\ref{sec:methodology}. Markers
        display the mean over three random seeds.}
    \label{fig:depth}
    \begin{center}
      \includegraphics[width=0.45\textwidth]{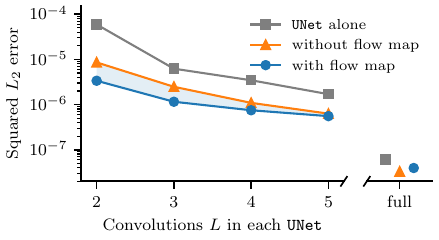}
    \end{center}
  \end{figure}

  Our network learns this simplified problem almost exactly,
    and as such, we view a small network here as a proxy for a network
    applied to more complex dynamics, and are most interested in what
    it gains from the flow map. With this in mind, we also train networks where each \texttt{UNet} is reduced to $L$
    convolutions of kernel size $3$ (see Appendix \ref{a:training}),
    such that each output of a \texttt{UNet} only depends on the
    $2L+1$ nearest DOF, compared to $19$ for the \texttt{UNet} above.
    In Figure \ref{fig:depth} we compare \eqref{eqn:nn} with the same
    network without the flow map, that is $\operatorname{UNet}(\vec{x}_c) +
    \operatorname{UNet2}(\operatorname{UNet}(\vec{x}_c))$, which has the same
    number of parameters, and with \texttt{UNet} alone as a baseline.
    Further, we display the networks with the \texttt{UNet} above
    separately (marked ``full'').

  We observe that, for $L = 2$ and $3$, the flow map lowers
    the squared error on every validation run and more than halves its
    average. We notice that the gain decreases as $L$ increases, which
    suggests that as a network sees more DOF it is able to learn more
    of the fine dynamics itself. In fact, for the full network the flow
    map increases the error by approximately $20\%$, although both
    networks improve upon \texttt{UNet} alone. As our training and
    validation runs come from the same family, the full network is
    likely able to fit all of their behaviour, whereas a small network is
    not. In Appendix
    \ref{a:training} we further verify that this gain is due to the fine
    dynamics in the matrix operator $A$.
    \vspace{-0.3cm}
  \vspace{-1cm}

  \section{Conclusion}

  In this paper, we introduced the problem of LAMs and outlined how
  neural networks can be used to improve regional weather
  forecasts. We considered a simplified set-up, and designed a neural
  interpolation operator to map dynamics between differing grid
  scales. We found that these operators can be
  learnt accurately. Further, building the
  flow of the fine dynamics into the network more than halves the
  square of the $L_2$ error of small networks. Before
  these methods can be effectively utilised for meteorological
  applications, our techniques need to be
  generalised to be applied in
  $2$D space and for higher order finite element methods, which
  requires graph neural networks and is ongoing work.

  \section*{Acknowledgements}
  This work was supported by the European Union’s Horizon Europe
  research and innovation programme under the Marie Sk{\l}odowska-Curie
  grant agreement No 101108679 (JJ) and by the UKRI Turing AI
  Acceleration Fellowship EP/V025295/2 (OJS).

  \bibliographystyle{abbrv}
  \bibliography{../lams}

  \appendix
  \renewcommand{\appendixname}{Appendix}
  \section{Discrete dynamical core} \label{a:fe}
  In this appendix, we shall discuss how the SWE \eqref{eqn:swe} is
  discretised to generate our coarse and fine scale data. Spatially, we discretise with
  finite elements and temporally with finite differences. We
  subdivide the periodic interval $[0,3)$ into the elements $I_m =
  (x_{m-1},x_{m})$, $m=1,\dots,M$, where $0=:x_0<x_1<\dots<x_M:=3$. We define the continuous finite element space
  $\dpoly{1}$ to be the space of functions which are linear over each
  $I_m$ and are globally continuous. We further define the
  discontinuous finite element space $\dpoly{0}$ as the space of functions which
  are constant over each $I_m$. With these spaces in mind, we
  spatially discretise \eqref{eqn:swe} by seeking some $U,V \in
  \dpoly{1}$ and $P\in\dpoly{0}$ such that
  \begin{equation}
    \begin{split}
      \inner{U_t - f V, \phi} - \inner{P, \phi_x} & = 0
                                                    \qquad
                                                    \forall \phi \in \dpoly{1}
      \\
      \inner{V_t + f U, \psi} & = 0
                                \qquad
                                \forall \psi \in \dpoly{1}
      \\
      \inner{P_t + g U_x, \chi} & = 0
                                  \qquad
                                  \forall \chi \in \dpoly{0}
                                  ,
    \end{split}
  \end{equation}
  where
  $\inner{\cdot,\cdot}$ denotes the spatial $L_2$ inner product over
  the periodic spatial domain. We note that this spatial
  discretisation is energy conserving, i.e.,
  \begin{equation}
    \ddt\bs{
      g \inner{U,U}
      + g \inner{V,V}
      + \inner{P,P}
    }
    =
    0
    .
  \end{equation}
  Temporally, we discretise with the implicit midpoint rule. Let
  $U^0,V^0\in\dpoly{1}$ and $P^0\in\dpoly{0}$ be given (by the
  interpolant of the initial data into the finite element space), then
  we seek $U^1,V^1\in\dpoly{1}$ and $P^1\in\dpoly{0}$ such that
  \begin{equation}
    \begin{split}
      \inner{\frac{U^1-U^0}{\tau} - f V^{\frac12}, \phi}
      - \inner{P^{\frac12},\phi_x} & = 0
                                     \qquad
                                     \forall \phi \in \dpoly{1} \\
      \inner{\frac{V^1-V^0}{\tau} + f U^{\frac12}, \psi} & = 0
                                                            \qquad
                                                            \forall \psi \in \dpoly{1} \\
      \inner{\frac{P^1-P^0}{\tau} + g U^{\frac12}_x, \chi} & = 0
                                                             \qquad
                                                             \forall \chi \in \dpoly{0}
                                                             ,
    \end{split}
  \end{equation}
  where $\tau$ is the time step, $U^{\frac12}:=\frac12\bc{U^1+U^0}$, and $V^{\frac12},
  P^{\frac12}$ are defined similarly. This fully discrete scheme
  has been designed such that energy is still preserved, i.e., that
  \begin{equation}
    \begin{split}
      g \inner{U^1,U^1} &
      + g \inner{V^1,V^1}
      + \inner{P^1,P^1}
      = \\
      & g \inner{U^0,U^0}
      + g \inner{V^0,V^0}
      + \inner{P^0,P^0}
      ,
    \end{split}
  \end{equation}
  and guarantees numerical stability over long time by preserving a
  modified $L_2$ norm of the solution.

  \section{Training details} \label{a:training}
    Here we describe how we train the networks of
    \S\ref{sec:implementation}, and give the details of the further
    experiments discussed alongside Figure \ref{fig:depth}. Throughout,
    we measure the error at a given time level by the square of the
    $L_2$ error in the finite element space, that is
    $\bc{\vec{y} - \vec{y}_f}^T \mathcal{M} \bc{\vec{y} - \vec{y}_f}$, where
    $\vec{y}$ and $\vec{y}_f$ denote the output of the network and the
    fine solution, and $\mathcal{M}$ is the mass matrix of the fine finite element
    space. We note that if we instead considered the mean squared error,
    our errors would be smaller in magnitude, but here we choose the
    $L_2$ error, since the mean squared error is not a trustworthy measure of error
    for a finite element function. At fixed resolution, our error may
    be thought of as
    a reweighted mean squared error. In Table \ref{tab:loss} and Figure
    \ref{fig:depth}, this error is averaged over the time levels and
    validation runs.

    We train each network with the Adam optimiser over $300$ epochs,
    with a batch size of $16$ and a learning rate which is updated after
    every batch. In particular, the learning rate rises linearly from
    zero to $10^{-3}$ over the first $5$ epochs, and then follows a
    cosine decay to zero, and we report each network after the final
    epoch. The loss is the square of the $L_2$ error, averaged over the
    time levels and over the batch. For $\sigma > 0$, we add a penalty
    to the loss as follows: For an input $\vec{x}_c$, we extract the
    value of the energy at the initial time and penalise the difference
    between the initial energy and the energy of $\NN(\vec{x}_c)$ in
    absolute value. This penalty is averaged over time and over the
    batch, and is multiplied by $\sigma$. With this penalty, the
    relative deviation of the energy of the output from the initial
    energy is on average approximately $3 \cdot 10^{-5}$ for each
    $\sigma > 0$, compared to $4 \cdot 10^{-5}$ for $\sigma = 0$.
    Further, we divide the loss by
    the square of the $L_2$ error of the coarse solution interpolated
    onto the fine mesh, averaged over the training data, such that it is
    of order one at the beginning of training. For $\sigma = 0$, the
    square of the $L_2$ error is over five orders of magnitude smaller
    than that of the coarse solution interpolated onto the fine mesh,
    with a relative $L_2$ error below $0.05\%$ compared to almost
    $20\%$.

    The convolutions of each \texttt{UNet} and \texttt{UNet2} act on
    data of order one. That is to say, for each component of the
    solution, we divide their input by the standard deviation of this
    component over the coarse training data, and multiply their output
    by the standard deviation over the training data of the difference
    between the fine and coarse solutions in this component. All
    networks in \texttt{UNet} and \texttt{UNet2} begin with the same weights, given by the default
    initialisation of PyTorch, and we train every network for the random
    seeds $1$, $2$ and $7$, reporting the mean and population standard
    deviation over these seeds. In Figure \ref{fig:depth}, these
    standard deviations are smaller than the markers.

    In Figure \ref{fig:network}, we display the structure of
    \eqref{eqn:nn}. All convolutions in this work are padded periodically, such that the
    number of DOF is preserved. For the smaller networks of Figure
    \ref{fig:depth}, we replace the convolutions of each \texttt{UNet}
    and \texttt{UNet2} by $L$ convolutions of kernel size $3$ with a
    ReLU between consecutive convolutions, where the first maps the $N$
    input channels to $s_1$ channels, the last maps $s_1$ channels back
    to $N$, and the $L-2$ in between map $s_1$ channels to $s_1$
    channels. The network without the flow map is \eqref{eqn:nn} with
    the loop over the time levels removed from \texttt{LearnFlow}. As
    $A$ contains no trainable parameters, this network has the same parameters as
    \eqref{eqn:nn}, of which there are approximately $3 \cdot 10^4$ for
    $L = 2$ and $4 \cdot 10^5$ for $L = 5$, compared to $2 \cdot 10^7$
    for the full network, and \texttt{UNet} alone has half as many.

    \begin{figure}[H]
    \centering
    \caption{The structure of our network \eqref{eqn:nn}. We represent
      the convolutions which act on each component of the solution by a
      white box, labelled by this component, and the flow map, which has
      no trainable parameters, by the shaded box. Note that no two white
      boxes share weights, that only \texttt{UNet} has a skip connection,
      and that $A^i$, which maps the first time level to each time level
      $i$, acts on all components at once.}
    \label{fig:network}
    \input{figures/network.tex}
  \end{figure}

  To verify that the gain from the flow map is due to the fine
  dynamics in $A$, we replace $A$ in \texttt{LearnFlow} by the
  identity, or by $S Q S^{-1}$, where $Q$ is a fixed random orthogonal
  matrix and $S$ is the diagonal matrix of the standard deviations by
  which we scale the input of each \texttt{UNet}. Like $A$, the latter
  mixes all DOF and components, but contains no dynamics. For $L = 2$
  and $3$, with the identity the error is then $4.6$ and $3.6$ times
  larger than that of \eqref{eqn:nn}, and with $S Q S^{-1}$ it is $5.4$
  and $4.8$ times larger, compared to $2.5$ and $2.1$ times larger without
  the flow map. In each case, \eqref{eqn:nn} has a smaller error on
  every validation run.

\end{multicols}
\end{document}

%% file: figures/mesh.tex
\begin{tikzpicture}[font=\footnotesize,
    declare function={
      fx(\a,\b) = \R*(-sin(\az) + cos(\az)*tan(\a))/sqrt(1 + tan(\a)^2 + tan(\b)^2);
      fy(\a,\b) = \R*(-sin(\el)*cos(\az) - sin(\el)*sin(\az)*tan(\a) + cos(\el)*tan(\b))
                  /sqrt(1 + tan(\a)^2 + tan(\b)^2);
    },
    coarse/.style={blue!55!black, line width=0.3pt},
    panel/.style={blue!55!black, line width=0.6pt},
    fine/.style={black!70, line width=0.12pt}]
  \def\R{1.7}   
  \def\az{7}   
  \def\el{15}   
  \def\n{6}     
  \def\ia{1} \def\ib{5} \def\ja{1} \def\jb{5} \def\m{2}

  \pgfmathsetmacro{\dx}{cos(\el)*cos(\az)} \pgfmathsetmacro{\dy}{cos(\el)*sin(\az)}
  \pgfmathsetmacro{\dz}{sin(\el)}
  \pgfmathsetmacro{\rx}{-sin(\az)} \pgfmathsetmacro{\ry}{cos(\az)}
  \pgfmathsetmacro{\ux}{-sin(\el)*cos(\az)} \pgfmathsetmacro{\uy}{-sin(\el)*sin(\az)}
  \pgfmathsetmacro{\uz}{cos(\el)}

  \shade[inner color=white, outer color=blue!30] (0,0) circle (\R);

  \foreach \cx/\cy/\cz/\px/\py/\pz/\qx/\qy/\qz in
      {1/0/0/0/1/0/0/0/1, -1/0/0/0/-1/0/0/0/1, 0/1/0/-1/0/0/0/0/1,
       0/-1/0/1/0/0/0/0/1, 0/0/1/0/1/0/-1/0/0, 0/0/-1/0/1/0/1/0/0} {
    \foreach \k in {0,...,\n} {
      \pgfmathsetmacro{\A}{tan(-45 + 90*\k/\n)}
      \pgfmathtruncatemacro{\edge}{or(\k == 0, \k == \n) ? 1 : 0}
      \foreach \fx/\fy/\fz/\vx/\vy/\vz in {\px/\py/\pz/\qx/\qy/\qz, \qx/\qy/\qz/\px/\py/\pz} {
        \pgfmathsetmacro{\zo}{\dx*(\cx + \A*\fx) + \dy*(\cy + \A*\fy) + \dz*(\cz + \A*\fz)}
        \pgfmathsetmacro{\zv}{\dx*\vx + \dy*\vy + \dz*\vz}
        \pgfmathsetmacro{\azv}{abs(\zv)}
        \ifdim\azv pt>0.001pt \pgfmathsetmacro{\sz}{-\zo/\zv}\else \def\sz{0}\fi
        \pgfmathsetmacro{\slo}{\zv > 0.001 ? max(-1, \sz) : -1}
        \pgfmathsetmacro{\shi}{\zv < -0.001 ? min(1, \sz) : 1}
        \pgfmathtruncatemacro{\ok}{and(\slo < \shi, or(\azv > 0.001, \zo > 0)) ? 1 : 0}
        \ifnum\ok=1
          \pgfmathsetmacro{\xo}{\rx*(\cx + \A*\fx) + \ry*(\cy + \A*\fy)}
          \pgfmathsetmacro{\xv}{\rx*\vx + \ry*\vy}
          \pgfmathsetmacro{\yo}{\ux*(\cx + \A*\fx) + \uy*(\cy + \A*\fy) + \uz*(\cz + \A*\fz)}
          \pgfmathsetmacro{\yv}{\ux*\vx + \uy*\vy + \uz*\vz}
          \pgfmathsetmacro{\tlo}{atan(\slo)} \pgfmathsetmacro{\thi}{atan(\shi)}
          \ifnum\edge=1 \def\sty{panel}\else \def\sty{coarse}\fi
          \draw[\sty]
            plot[domain=\tlo:\thi, samples=16, smooth]
            ({\R*(\xo + tan(\x)*\xv)/sqrt(1 + \A*\A + tan(\x)^2)},
             {\R*(\yo + tan(\x)*\yv)/sqrt(1 + \A*\A + tan(\x)^2)});
        \fi
      }
    }
  }

  \pgfmathsetmacro{\aa}{-45 + 90*\ia/\n} \pgfmathsetmacro{\ab}{-45 + 90*\ib/\n}
  \pgfmathsetmacro{\ba}{-45 + 90*\ja/\n} \pgfmathsetmacro{\bb}{-45 + 90*\jb/\n}
  \pgfmathsetmacro{\la}{-45 + 90*(\ia + \m/4)/\n} \pgfmathsetmacro{\lb}{-45 + 90*(\ib - \m/4)/\n}
  \pgfmathsetmacro{\ma}{-45 + 90*(\ja + \m/4)/\n} \pgfmathsetmacro{\mb}{-45 + 90*(\jb - \m/4)/\n}
  \fill[orange!60, opacity=0.85] plot[domain=\aa:\ab] ({fx(\x,\ba)}, {fy(\x,\ba)})
    -- plot[domain=\ba:\bb] ({fx(\ab,\x)}, {fy(\ab,\x)})
    -- plot[domain=\ab:\aa] ({fx(\x,\bb)}, {fy(\x,\bb)})
    -- plot[domain=\bb:\ba] ({fx(\aa,\x)}, {fy(\aa,\x)}) -- cycle;
  \fill[red!65, opacity=0.85] plot[domain=\la:\lb] ({fx(\x,\ma)}, {fy(\x,\ma)})
    -- plot[domain=\ma:\mb] ({fx(\lb,\x)}, {fy(\lb,\x)})
    -- plot[domain=\lb:\la] ({fx(\x,\mb)}, {fy(\x,\mb)})
    -- plot[domain=\mb:\ma] ({fx(\la,\x)}, {fy(\la,\x)}) -- cycle;
  \pgfmathtruncatemacro{\ka}{4*(\ib - \ia)} \pgfmathtruncatemacro{\kb}{4*(\jb - \ja)}
  \foreach \k in {0,...,\ka} {
    \pgfmathsetmacro{\t}{-45 + 90*(\ia + \k/4)/\n}
    \draw[fine] plot[domain=\ba:\bb, samples=12] ({fx(\t,\x)}, {fy(\t,\x)});
  }
  \foreach \k in {0,...,\kb} {
    \pgfmathsetmacro{\t}{-45 + 90*(\ja + \k/4)/\n}
    \draw[fine] plot[domain=\aa:\ab, samples=12] ({fx(\x,\t)}, {fy(\x,\t)});
  }
  \foreach \k in {\ia,...,\ib} {
    \pgfmathsetmacro{\t}{-45 + 90*\k/\n}
    \draw[coarse] plot[domain=\ba:\bb, samples=12] ({fx(\t,\x)}, {fy(\t,\x)});
  }
  \foreach \k in {\ja,...,\jb} {
    \pgfmathsetmacro{\t}{-45 + 90*\k/\n}
    \draw[coarse] plot[domain=\aa:\ab, samples=12] ({fx(\x,\t)}, {fy(\x,\t)});
  }
  \draw[line width=0.4pt] (0,0) circle (\R);

  \pgfmathsetmacro{\mam}{(\ma + \mb)/2}
  \pgfmathsetmacro{\lam}{\lb - 90/8/\n} \pgfmathsetmacro{\band}{\ab - 90*\m/8/\n}
  \fill ({fx(\lam,\mam + 8)}, {fy(\lam,\mam + 8)}) circle (0.7pt);
  \draw[line width=0.3pt] ({fx(\lam,\mam + 8)}, {fy(\lam,\mam + 8)}) -- (\R+0.35,0.95)
    node[right] {LAM};
  \fill ({fx(\band,\mam - 4)}, {fy(\band,\mam - 4)}) circle (0.7pt);
  \draw[line width=0.3pt] ({fx(\band,\mam - 4)}, {fy(\band,\mam - 4)}) -- (\R+0.35,0.15)
    node[right, align=left] {Blending/interface\\region};
  \fill ({fx(38,-38)}, {fy(38,-38)}) circle (0.7pt);
  \draw[line width=0.3pt] ({fx(38,-38)}, {fy(38,-38)}) -- (\R+0.35,-0.95)
    node[right] {Global model};
\end{tikzpicture}

%% file: figures/fes.tex
\begin{tikzpicture}[x=0.6cm, y=0.6cm, font=\footnotesize, black,
    dof/.style={cross out, draw, thick, minimum size=4.5pt, inner sep=0pt, outer sep=1pt}]
  \draw[semithick] (0,0) -- (10,0);
  \foreach \x/\l in {0/0, 5/\frac32, 10/3}
    \draw[semithick] (\x,0.15) -- (\x,-0.15) node[below] {$\l$};
  \foreach \x/\l in {2.5/\frac34, 7.5/\frac94}
    \draw (\x,0.08) -- (\x,-0.08) node[below] {$\l$};
  \foreach \x/\y in {0/1, 5/2, 10/1, 2.5/1.2, 7.5/0.8}
    \draw[densely dotted, black!45] (\x,\y) -- (\x,0);
  \draw[thick, blue!65!black] (0,1) -- (5,2) -- (10,1) node[pos=0.8, above] {$u$};
  \foreach \x/\y in {0/1, 5/2, 10/1} \node[dof, blue!65!black] at (\x,\y) {};
  \draw[thick, red] (0,1.2) -- (5,1.2) (5,0.8) -- (10,0.8) node[pos=0.75, below] {$p$};
  \foreach \x/\y in {2.5/1.2, 7.5/0.8} \node[dof, red] at (\x,\y) {};
\end{tikzpicture}

%% file: figures/network.tex
\begin{tikzpicture}[font=\footnotesize, black, >=stealth, line width=0.4pt,
    conv/.style={draw, minimum width=0.55cm, minimum height=0.36cm, inner sep=0pt},
    sum/.style={circle, draw, minimum size=0.24cm, inner sep=0pt,
      path picture={\draw (path picture bounding box.north) -- (path picture bounding box.south)
        (path picture bounding box.west) -- (path picture bounding box.east);}}]
  \node[anchor=east] (in) at (0.3,0) {$\vec{x}_c$};
  \foreach \c/\y in {u/0.45, v/0, p/-0.45} {
    \node[conv] (a\c) at (1.2,\y) {$\c$};
    \draw[->] (0.55,\y) -- (a\c.west);
    \draw (a\c.east) -- (1.75,\y);
  }
  \draw (in.east) -- (0.55,0) (0.55,-0.45) -- (0.55,0.45) (1.75,-0.45) -- (1.75,0.45);
  \node[sum] (s1) at (2.2,0) {};
  \draw[->] (1.75,0) -- (s1);
  \draw[->] (0.55,0) |- (1.2,-0.9) -| (s1.south);
  \fill (0.55,0) circle (0.9pt);
  \node[above] at (1.2,0.65) {\texttt{UNet}};
  \node[below] at (1.2,-0.9) {skip connection};
  \draw (s1.east) -- (2.65,0);
  \node[sum] (s2) at (5.5,0) {};
  \draw[->] (2.65,0) |- node[pos=0.75, above] {$\operatorname{UNet}(\vec{x}_c)$} (5.5,1.15) -- (s2.north);
  \node[draw, fill=orange!40, minimum width=0.6cm, minimum height=1.27cm] (flow) at (3.25,0) {$A^i$};
  \draw[->] (2.65,0) -- (flow.west);
  \fill (2.65,0) circle (0.9pt);
  \foreach \c/\y in {u/0.45, v/0, p/-0.45} {
    \node[conv] (b\c) at (4.45,\y) {$\c$};
    \draw[->] (3.55,\y) -- (b\c.west);
    \draw (b\c.east) -- (5.0,\y);
  }
  \draw (5.0,-0.45) -- (5.0,0.45);
  \draw[->] (5.0,0) -- (s2);
  \node[above] at (4.45,0.65) {\texttt{UNet2}};
  \draw (2.95,-0.8) -- (2.95,-0.9) -- (4.725,-0.9) -- (4.725,-0.8);
  \node[below] at (3.84,-0.9) {\texttt{LearnFlow}};
  \draw[->] (s2.east) -- (5.95,0) node[right] {$\NN(\vec{x}_c)$};
\end{tikzpicture}

%% file: lamkin.bbl
\begin{thebibliography}{10}

\bibitem{Berliner2003}
L.~M. Berliner.
\newblock Physical‐statistical modeling in geophysics.
\newblock {\em Journal of Geophysical Research: Atmospheres}, 108(D24):8776,
  Dec. 2003.

\bibitem{Bush2020}
M.~Bush, T.~Allen, C.~Bain, I.~Boutle, J.~Edwards, A.~Finnenkoetter,
  C.~Franklin, K.~Hanley, H.~Lean, A.~Lock, J.~Manners, M.~Mittermaier,
  C.~Morcrette, R.~North, J.~Petch, C.~Short, S.~Vosper, D.~Walters,
  S.~Webster, M.~Weeks, J.~Wilkinson, N.~Wood, and M.~Zerroukat.
\newblock The first {Met Office Unified Model–JULES Regional Atmosphere and
  Land} configuration, {RAL1}.
\newblock {\em Geoscientific Model Development}, 13(4):1999--2029, Apr. 2020.

\bibitem{Bush2023}
M.~Bush, I.~Boutle, J.~Edwards, A.~Finnenkoetter, C.~Franklin, K.~Hanley,
  A.~Jayakumar, H.~Lewis, A.~Lock, M.~Mittermaier, S.~Mohandas, R.~North,
  A.~Porson, B.~Roux, S.~Webster, and M.~Weeks.
\newblock The second {Met Office Unified Model–JULES Regional Atmosphere and
  Land} configuration, {RAL2}.
\newblock {\em Geoscientific Model Development}, 16(6):1713--1734, Mar. 2023.

\bibitem{CardosoBihlo2025}
E.~Cardoso-Bihlo and A.~Bihlo.
\newblock Exactly conservative physics-informed neural networks and deep
  operator networks for dynamical systems.
\newblock {\em Neural Networks}, 181:106826, Jan. 2025.

\bibitem{Celledoni2021}
E.~Celledoni, M.~J. Ehrhardt, C.~Etmann, R.~I. McLachlan, B.~Owren, C.-B.
  Schönlieb, and F.~Sherry.
\newblock Structure-preserving deep learning.
\newblock {\em European Journal of Applied Mathematics}, 32(5):888--936, Oct.
  2021.

\bibitem{self:pixels}
E.~Celledoni, J.~Jackaman, D.~Murari, and B.~Owren.
\newblock Predictions based on pixel data: Insights from {PDEs} and finite
  differences.
\newblock {\em Journal of Computational Physics}, 538:114166, Oct. 2025.

\bibitem{Celledoni2023}
E.~Celledoni, D.~Murari, B.~Owren, C.-B. Schönlieb, and F.~Sherry.
\newblock Dynamical systems–based neural networks.
\newblock {\em SIAM Journal on Scientific Computing}, 45(6):A3071--A3094, Dec.
  2023.

\bibitem{Cotter2023}
C.~J. Cotter.
\newblock Compatible finite element methods for geophysical fluid dynamics.
\newblock {\em Acta Numerica}, 32:291--393, May 2023.

\bibitem{Cotter2014}
C.~J. Cotter and J.~Thuburn.
\newblock A finite element exterior calculus framework for the rotating
  shallow-water equations.
\newblock {\em Journal of Computational Physics}, 257:1506--1526, Jan. 2014.

\bibitem{Danieli2022}
F.~Danieli, B.~S. Southworth, and A.~J. Wathen.
\newblock Space-time block preconditioning for incompressible flow.
\newblock {\em SIAM Journal on Scientific Computing}, 44(1):A337--A363, Feb.
  2022.

\bibitem{David2023}
M.~David and F.~Méhats.
\newblock Symplectic learning for {H}amiltonian neural networks.
\newblock {\em Journal of Computational Physics}, 494:112495, Dec. 2023.

\bibitem{Dong2017}
B.~Dong, Q.~Jiang, and Z.~Shen.
\newblock Image restoration: Wavelet frame shrinkage, nonlinear evolution
  {PDEs}, and beyond.
\newblock {\em Multiscale Modeling \& Simulation}, 15(1):606--660, Jan. 2017.

\bibitem{Dong2016}
C.~Dong, C.~C. Loy, K.~He, and X.~Tang.
\newblock Image super-resolution using deep convolutional networks.
\newblock {\em IEEE Transactions on Pattern Analysis and Machine Intelligence},
  38(2):295--307, Feb. 2016.

\bibitem{Fisher2008}
M.~Fisher, J.~Nocedal, Y.~Trémolet, and S.~J. Wright.
\newblock Data assimilation in weather forecasting: a case study in
  {PDE}-constrained optimization.
\newblock {\em Optimization and Engineering}, 10(3):409--426, Sept. 2009.

\bibitem{Firedrake}
D.~A. Ham, P.~H.~J. Kelly, L.~Mitchell, C.~J. Cotter, R.~C. Kirby, K.~Sagiyama,
  N.~Bouziani, S.~Vorderwuelbecke, T.~J. Gregory, J.~Betteridge, D.~R. Shapero,
  R.~W. Nixon-Hill, C.~J. Ward, P.~E. Farrell, P.~D. Brubeck, I.~Marsden, T.~H.
  Gibson, M.~Homolya, T.~Sun, A.~T.~T. McRae, F.~Luporini, A.~Gregory,
  M.~Lange, S.~W. Funke, F.~Rathgeber, G.-T. Bercea, and G.~R. Markall.
\newblock {\em Firedrake User Manual}.
\newblock Imperial College London and University of Oxford and Baylor
  University and University of Washington, first edition, May 2023.

\bibitem{He2016}
K.~He, X.~Zhang, S.~Ren, and J.~Sun.
\newblock Deep residual learning for image recognition.
\newblock In {\em 2016 IEEE Conference on Computer Vision and Pattern
  Recognition (CVPR)}, pages 770--778, 2016.

\bibitem{self:code}
J.~Jackaman.
\newblock Implementation of ``{I}mproving regional weather forecasts with
  neural interpolation''.
\newblock https://doi.org/10.5281/zenodo.14888898, 2026.

\bibitem{self:rewaves}
J.~Jackaman and S.~MacLachlan.
\newblock Space-time waveform relaxation multigrid for {N}avier--{S}tokes.
\newblock {\em SIAM Journal on Scientific Computing}, 48(5):S555--S585, Oct.
  2026.

\bibitem{Jagtap2020}
A.~D. Jagtap, E.~Kharazmi, and G.~E. Karniadakis.
\newblock Conservative physics-informed neural networks on discrete domains for
  conservation laws: Applications to forward and inverse problems.
\newblock {\em Computer Methods in Applied Mechanics and Engineering},
  365:113028, June 2020.

\bibitem{Kent2023}
J.~Kent, T.~Melvin, and G.~A. Wimmer.
\newblock A mixed finite-element discretisation of the shallow-water equations.
\newblock {\em Geoscientific Model Development}, 16(4):1265--1276, Feb. 2023.

\bibitem{Lam2023}
R.~Lam, A.~Sanchez-Gonzalez, M.~Willson, P.~Wirnsberger, M.~Fortunato, F.~Alet,
  S.~Ravuri, T.~Ewalds, Z.~Eaton-Rosen, W.~Hu, A.~Merose, S.~Hoyer, G.~Holland,
  O.~Vinyals, J.~Stott, A.~Pritzel, S.~Mohamed, and P.~Battaglia.
\newblock Learning skillful medium-range global weather forecasting.
\newblock {\em Science}, 382(6677):1416--1421, Dec. 2023.

\bibitem{Lohmann2023}
C.~Lohmann and S.~Turek.
\newblock On the design of global-in-time {N}ewton-multigrid-pressure {S}chur
  complement solvers for incompressible flow problems.
\newblock {\em Journal of Mathematical Fluid Mechanics}, 25(3):64, Aug. 2023.

\bibitem{Long2015}
J.~Long, E.~Shelhamer, and T.~Darrell.
\newblock Fully convolutional networks for semantic segmentation.
\newblock In {\em 2015 IEEE Conference on Computer Vision and Pattern
  Recognition (CVPR)}, pages 3431--3440, 2015.

\bibitem{Maynard2020}
C.~Maynard, T.~Melvin, and E.~H. Müller.
\newblock Multigrid preconditioners for the mixed finite element dynamical core
  of the {LFRic} atmospheric model.
\newblock {\em Quarterly Journal of the Royal Meteorological Society},
  146(733):3917--3936, Oct. 2020.

\bibitem{Melvin2018}
T.~Melvin.
\newblock Dispersion analysis of the {$P_n$}--{$P_{n-1}^{DG}$} mixed finite
  element pair for atmospheric modelling.
\newblock {\em Journal of Computational Physics}, 355:342--365, Feb. 2018.

\bibitem{Melvin2024}
T.~Melvin, B.~Shipway, N.~Wood, T.~Benacchio, T.~Bendall, I.~Boutle, A.~Brown,
  C.~Johnson, J.~Kent, S.~Pring, C.~Smith, M.~Zerroukat, C.~Cotter, and
  J.~Thuburn.
\newblock A mixed finite‐element, finite‐volume, semi‐implicit
  discretisation for atmospheric dynamics: Spherical geometry.
\newblock {\em Quarterly Journal of the Royal Meteorological Society},
  150(764):4252--4269, Oct. 2024.

\bibitem{Muller2014}
E.~H. Müller and R.~Scheichl.
\newblock Massively parallel solvers for elliptic partial differential
  equations in numerical weather and climate prediction.
\newblock {\em Quarterly Journal of the Royal Meteorological Society},
  140(685):2608--2624, Oct. 2014.

\bibitem{Navon}
I.~M. Navon.
\newblock Data assimilation for numerical weather prediction: A review.
\newblock In S.~K. Park and L.~Xu, editors, {\em Data Assimilation for
  Atmospheric, Oceanic and Hydrologic Applications}, pages 21--65. Springer
  Berlin Heidelberg, 2009.

\bibitem{PyTorch}
A.~Paszke, S.~Gross, F.~Massa, A.~Lerer, J.~Bradbury, G.~Chanan, T.~Killeen,
  Z.~Lin, N.~Gimelshein, L.~Antiga, A.~Desmaison, A.~Kopf, E.~Yang, Z.~DeVito,
  M.~Raison, A.~Tejani, S.~Chilamkurthy, B.~Steiner, L.~Fang, J.~Bai, and
  S.~Chintala.
\newblock {PyTorch}: An imperative style, high-performance deep learning
  library.
\newblock In {\em Advances in Neural Information Processing Systems 32}, pages
  8024--8035. Curran Associates, Inc., 2019.

\bibitem{Pratt2006}
W.~K. Pratt.
\newblock {\em Digital Image Processing: {PIKS} Scientific Inside}.
\newblock Wiley, fourth edition, June 2006.

\bibitem{Rabier2005}
F.~Rabier.
\newblock Overview of global data assimilation developments in numerical
  weather‐prediction centres.
\newblock {\em Quarterly Journal of the Royal Meteorological Society},
  131(613):3215--3233, Oct. 2005.

\bibitem{Siddique2021}
N.~Siddique, S.~Paheding, C.~P. Elkin, and V.~Devabhaktuni.
\newblock {U-Net} and its variants for medical image segmentation: A review of
  theory and applications.
\newblock {\em IEEE Access}, 9:82031--82057, 2021.

\bibitem{Staniforth2013}
A.~Staniforth, T.~Melvin, and C.~Cotter.
\newblock Analysis of a mixed finite‐element pair proposed for an atmospheric
  dynamical core.
\newblock {\em Quarterly Journal of the Royal Meteorological Society},
  139(674):1239--1254, July 2013.

\bibitem{Straka2009}
J.~M. Straka.
\newblock {\em Cloud and Precipitation Microphysics: Principles and
  Parameterizations}.
\newblock Cambridge University Press, 2009.

\bibitem{Ullrich2017}
P.~A. Ullrich, C.~Jablonowski, J.~Kent, P.~H. Lauritzen, R.~Nair, K.~A. Reed,
  C.~M. Zarzycki, D.~M. Hall, D.~Dazlich, R.~Heikes, C.~Konor, D.~Randall,
  T.~Dubos, Y.~Meurdesoif, X.~Chen, L.~Harris, C.~Kühnlein, V.~Lee,
  A.~Qaddouri, C.~Girard, M.~Giorgetta, D.~Reinert, J.~Klemp, S.-H. Park,
  W.~Skamarock, H.~Miura, T.~Ohno, R.~Yoshida, R.~Walko, A.~Reinecke, and
  K.~Viner.
\newblock {DCMIP2016}: a review of non-hydrostatic dynamical core design and
  intercomparison of participating models.
\newblock {\em Geoscientific Model Development}, 10(12):4477--4509, Dec. 2017.

\bibitem{Wang2021}
Z.~Wang, J.~Chen, and S.~C.~H. Hoi.
\newblock Deep learning for image super-resolution: A survey.
\newblock {\em IEEE Transactions on Pattern Analysis and Machine Intelligence},
  43(10):3365--3387, Oct. 2021.

\bibitem{Wesseling2001}
P.~Wesseling and C.~W. Oosterlee.
\newblock Geometric multigrid with applications to computational fluid
  dynamics.
\newblock {\em Journal of Computational and Applied Mathematics},
  128(1–2):311--334, Mar. 2001.

\end{thebibliography}
